\documentclass[conference]{IEEEtran}
\IEEEoverridecommandlockouts
\usepackage{cite}
\usepackage{amsmath,amssymb,amsfonts}
\usepackage{algorithm}
\usepackage{algorithmic}
\usepackage[utf8]{inputenc}
\usepackage{multirow}
\usepackage{array}
\usepackage{graphicx}
\usepackage{lipsum}
\usepackage{subcaption}
\usepackage{textcomp}
\usepackage{xcolor}
\usepackage{amsmath}
\usepackage[left=0.625in, right=0.625in, top=0.75in, bottom=1in, footskip=0.25in]{geometry}
\usepackage{tikz}
\usepackage{amssymb}
\usepackage{graphicx}
\usepackage{cite}
\usepackage{bm}
\usepackage{dsfont}
\usepackage{graphics} 
\usepackage{epsfig} 
\usepackage{stackrel}
\usepackage{geometry}{}
\usepackage{float}
\usepackage{url}
\usepackage[colorlinks,citecolor=blue]{hyperref}

\usepackage{ifthen}

\newboolean{showcomments}
\setboolean{showcomments}{true}
\ifthenelse{\boolean{showcomments}}
{ \newcommand{\mynote}[3]{
    \protect\fbox{\sffamily\scriptsize#1}
    {\small$\blacktriangleright$\textsf{\emph{\color{#3}{#2}}}$\blacktriangleleft$}}}
{ \newcommand{\mynote}[3]{}}

\def\BibTeX{{\rm B\kern-.05em{\sc i\kern-.025em b}\kern-.08em
    T\kern-.1667em\lower.7ex\hbox{E}\kern-.125emX}}
\begin{document}

\title{\textbf{OPTILOD:} \textbf{O}ptimal Beacon \textbf{P}lacemen\textbf{t} for High-Accuracy \textbf{I}ndoor \textbf{Lo}calization of \textbf{D}rones}

\author{\IEEEauthorblockN{Alireza Famili, Angelos Stavrou, Haining Wang, Jung-Min (Jerry) Park}
	\IEEEauthorblockA{\textit{Department of Electrical and Computer Engineering, Virginia Tech} \\
		\{afamili, angelos, hnw, jungmin\}@vt.edu}
}

\maketitle

\begin{abstract}
For many applications, drones are required to operate entirely or partially autonomously. To fly completely or partially on their own, drones need access to location services to get navigation commands. While using the Global Positioning System (GPS) is an obvious choice, GPS is not always available, can be spoofed or jammed, and is highly error-prone for indoor and underground environments. The ranging method using beacons is one of the popular methods for localization, specially for indoor environments. In general, localization error in this class is due to two factors: the ranging error and the error induced by the relative geometry between the beacons and the target object to localize. This paper proposes OPTILOD (Optimal Beacon Placement for High-Accuracy Indoor Localization of Drones), an optimization algorithm for the optimal placement of beacons deployed in three-dimensional indoor environments. OPTILOD leverages advances in Evolutionary Algorithms to compute the minimum number of beacons and their optimal placement to minimize the localization error. These problems belong to the Mixed Integer Programming (MIP) class and are both considered NP-Hard. Despite that, OPTILOD can provide multiple optimal beacon configurations that minimize the localization error and the number of deployed beacons concurrently and time efficiently. 
\end{abstract}


\section{Introduction} \label{s:Introduction}
\noindent Over the last few years, there has been a renewed interest and investment in drone technologies for both indoors and outdoor applications. Indeed, there are a variety of indoor drone applications available today that range from recreational to life-saving. Some examples where drones have made a significant impact include reconnaissance inside nuclear power plants, assisting firefighters in locating people trapped inside burning buildings, and security monitoring inside large warehouses~\cite{wawrla2019applications,DronePower2021}.

In order to successfully perform their mission, drones should have entirely or partially autonomous flying capability in the majority of the above applications. To allow these self-piloted flights, the drone must first be able to constantly localize itself, after which a navigation command signal is produced and sent to the drone's controller unit based on the current location and ultimate destination. Drones can easily use the GPS signal for self-localization in outdoor settings, but this is not possible in areas where GPS signals are not accessible, such as indoor environments.

There is a plethora of localization approaches including vision-based~\cite{low_cost_solution, 6_Dimensional, Survey_UAV_navigation_GPS_denied}, using cellular networks~\cite{Signal_of_Opportunity}, and ranging-based using RF signals~\cite{Freq_Hopping_WiFi} or acoustic signals~\cite{ROLATIN} to name a few. In all the ranging-based techniques, localization is conducted using the processing of received signals. Thus, for all ranging-based techniques, there exist a number of transmitters and receivers installed on-board the drone and at known locations in the surrounding area. Localization is performed by measuring information derived from the transmitted signal between the sensors on-board the drone and ones (also called beacons) placed in the surrounding environment.

In this paper, we propose an optimization scheme that computes the minimal number of beacons required to provide localization capability for the entire indoor environment. Our approach, called \emph{OPTILOD: Optimal Beacon Placement for High-Accuracy Indoor Localization of Drones}, is also able to determine the optimal placement of indoor beacons to mitigate the localization error. In the case of indoor beacons, the localization error is due to the relative geometry between the transmitter beacons and the receiver. To achieve localization, OPTILOD employs ultrasonic acoustic-based signals. We posit that acoustic signals have some benefits over RF signals in terms of localization. Indeed, acoustic signals have a much slower propagation speed than RF signals, which allows for better localization accuracy. Furthermore, RF signals are not suitable for indoor ranging as their signals can pass through walls, ceilings, and other man-made obstacles. This results in increased interference which negatively impacts the precision of the localization. Moreover, to avoid interference from human-generated or drone-generated propeller noise, OPTILOD employs high-frequency acoustic signals known as \emph{ultrasounds}.

OPTILOD focuses on the placement of beacons in the indoor environment to reduce localization errors. For the system to work, we need to guarantee signal coverage from at least four beacons at any point in the designated indoor flight area. In addition, a secondary optimization goal for the beacon placement is to reduce the localization error induced by the relative geometry between transmitter beacons and the receiver on-board the drone.
The main contributions of our work can be summarized as follows.

$\bullet$ We propose OPTILOD, an optimization algorithm for the optimal placement of beacons in three-dimensional indoor environments. 

$\bullet$ OPTILOD computes the \emph{minimum} number of beacons required to provide full coverage for any indoor area regardless of its dimensions, thereby reducing deployment costs.

$\bullet$ OPTILOD generates the optimal beacon placement that achieves both full coverage and minimizes the localization error induced by the relative geometry between the transmitters and the receivers.

$\bullet$ OPTILOD is the first to design an evolutionary algorithm to formulate these two NP-Hard beacon placement problems as a dual optimization objective and generate solutions in tractable time.

$\bullet$ We evaluate OPTILOD using comprehensive simulations and realistic environmental indoor parameters. Our results show that OPTILOD generates beacon placements that are either optimal or within one beacon difference to the theoretical optimal number of beacons while minimizing the localization error across the entire indoor area.

The rest of this paper is organized as follows. The next section reviews the related work, followed by a brief overview of how OPTILOD renders localization in the absence of GPS~\ref{s:Localization}. Then, in sections~\ref{s:Problem_Formulation} and~\ref{s:Algorithm_Design}, we detail the core contributions for this paper and include calculating the minimum number of beacons and their optimal placement to fully cover an indoor environment for ranging-based three-dimensional localization. 
Section~\ref{s:Results} provides simulation results for the proposed optimization scheme. Finally, we conclude our work in section~\ref{s:Conclusion}.

\section{Related Work} \label{s:Related Work}
Our work is related to the following research studies: (i) indoor localization, (ii) autonomous navigation of drones in the absence (or lack) of GPS signals, and (iii) optimal beacon placement.

Localizing a target in indoor environments and in the absence of the GPS signal has been a topic of interest. Ranging-based methods are of the most well-known approaches for indoor localization. In this category RF, acoustic, or ultrasound signals are getting deployed to find the distance between each of the beacons and the object to find the distance, then using the distance to different beacons to localize the object \cite{ROLATIN, CAT, MobiSys_Follow_Me_Drone, ToneTrack, Robust_Broadband}. 

For autonomous navigation of drones in the absence (or lack) of GPS signals, there are some well-known techniques that tackle the problem. For example, vision-based models using different visual techniques such as visual odometry (VO), simultaneous localization and mapping (SLAM), and optical flow \cite{low_cost_solution, 6_Dimensional, Survey_UAV_navigation_GPS_denied}. There are also a few research papers where they use deep neural networks in combination with visual techniques~\cite{Neural_Network} or use of LiDAR~\cite{LiDAR} for autonomous flying.

In terms of beacon placement optimization, this is a well-known topic to optimize the location of beacons for indoor localization purposes~\cite{CMU, Efficient_Beacon_Placement, CMU_paper, Novel_Beacon_Placement, BLE_Localization_Precision_Limits} and wireless network localization~\cite{Relative_location_estimation, Computational_Geometry_Framework, Network_Navigation_with_Scheduling}. There are two major issues here, first optimizing the number of beacons and their location to have full coverage for the entire indoor venue, and second optimizing the placement of beacons to minimize the localization error induced by the relative geometry between the target and beacons. For the first issue, the type of sensors plays an important role, because they have different coverage, e.g., if the system is based on low power Bluetooth sensors, the transmission would be omnidirectional and the distance and obstacles restrict coverage; however, if a system uses ultrasound-based sensors, then the beam angle of the sensors also puts restrictions on finding the number of sensors and their placement. Optimizing the number of required beacons is mostly used for finding the number of sensors for large indoor environments with different stories and rooms. On the other hand, after finding the number of beacons, the second optimization platform needs to be deployed to find the placement for sensors to minimize the localization error induced by the relative geometry between the transmitter and beacons. 

These two aforementioned optimizations-- optimizing the minimum number of beacons required for full coverage and optimizing the placement of beacons to mitigate the geometry induced error-- are both in the class of NP-Hard problems when it comes to the mobile object (drone) in the three-dimensional environments. The majority of the previous works focus on the problem of finding the minimum number of beacons required to fully cover an area, mostly in two dimensions, and without considering the importance of the relative geometry between the beacons and the receiver. A few research studies consider the geometry-induced error in their scheme, but mostly for two dimensions, which is not the case for a moving drone in space.

In this work, we propose a scheme to find the minimum number of beacons to provide full coverage in three-dimensional space. We also consider the placement of these beacons to reduce the geometry-induced error of localization for flying drones. 
Therefore, compared to prior work, our approach: (i) provides the solution for full coverage in three dimensions rather than just two dimensions, (ii) considers the importance of the relative geometry between the beacons and the object to localize, and most importantly (iii) provides the solution for a moving object (drone) rather than just a single static point.

\section{Localization in Absence of GPS} \label{s:Localization}
\subsection{Background}
There are different ways to perform localization in absence of GPS signals~\cite{low_cost_solution, 6_Dimensional, Survey_UAV_navigation_GPS_denied}. Some of the popular approaches use vision-based localization with optical sensors. 
In addition to vision-based localization, ranging-based localization is another well-established method. OPTILOD uses ranging-based techniques for localization primarily due to the cheap deployment cost, low computational complexity, and fast response times compared to vision-based methods. Due to the advantages of acoustic signals over RF for high-accuracy localization, OPTILOD employs an ultrasound acoustic signal for localization purposes.

Some of the well-known measurement methods for distance estimation include Angle Of Arrival (AOA), Time Of Arrival (TOA), Time Difference Of Arrival (TDOA), and Received Signal Strength (RSS). To perform location estimation, the options are angulation, lateration, and fingerprinting. AOA requires special antenna arrays and incurs high complexity calculations, which make the approach expensive in terms of cost and processing power. RSS and fingerprinting are too sensitive to real-time changes; hence, they are unreliable. OPTILOD uses trilateration techniques and the TOA of received ultrasound signals for localization. We assume that the ultrasound receiver is on-board the drone and the ultrasound transmitters are located at known locations in the room. We use the TOA of the received signal at the ultrasound receiver to estimate the distance between the receiver (drone) and the corresponding transmitter in the room. We use the following equation for distance estimation: $d = c \times t;$ where $d$ is the distance, $c$ is the speed of sound, and $t$ is the time of flight for the signal which is calculated based on the TOA.

\subsection{Three-dimensional Trilateration}
After successfully measuring the distance between an ultrasonic transmitter and the receiver, the next step is three-dimensional localizing the receiver (drone). For localizing an object in two dimensions using trilateration, we need to know the distance between the object and three locations (sources). Similarly, for three-dimensional localization, we need to know the distance between the object and at least four sources to be able to localize the object uniquely. Let's denote the distance between the receiver and $i$-th transmitter as $d_i$. Also, the position of the receiver is $[x \ y \ z]^T$ (which in fact is the position of drone), and the position of $i$-th transmitter denotes as $[x_i \ y_i \ z_i]^T$. Then using trilateration rules, we have:
\begin{eqnarray} \nonumber
	&(x_1-x)^2+(y_1-y)^2+(z_1-z)^2 = d_1^2\\ \nonumber
	&(x_2-x)^2+(y_2-y)^2+(z_2-z)^2 = d_2^2\\ \nonumber
	&\vdots\\
	&(x_n-x)^2+(y_n-y)^2+(z_n-z)^2 = d_n^2 
\end{eqnarray}
We can then simplify these quadratic equations and write them down in the form of $\textbf{A}\textbf{x} = \textbf{b}$ where $\textbf{A}$ and $\textbf{b}$ are equal to:
\begin{small}
	\begin{eqnarray}
		\textbf{A} & = & \begin{bmatrix}
			2(x_n-x_1) & 2(y_n-y_1) & 2(z_n-z_1) \\
			2(x_n-x_2) & 2(y_n-y_2) & 2(z_n-z_2)\\
			\vdots     & \vdots     & \vdots\\
			2(x_n-x_{n-1}) & 2(y_n-y_{n-1}) & 2(z_n-z_{n-1})\\
		\end{bmatrix},\nonumber \\
		\textbf{b} & = & \begin{bmatrix}
			d_1^2 - d_n^2 - x_1^2 -y_1^2 -z_1^2 + x_n^2 + y_n^2 + z_n^2 \\
			d_2^2 - d_n^2 - x_2^2 -y_2^2 -z_2^2 + x_n^2 + y_n^2 + z_2^2 \\
			\vdots \\
			d_{n-1}^2 - d_n^2 - x_{n-1}^2 -y_{n-1}^2 -z_{n-1}^2 + x_n^2 + y_n^2 + z_n^2
		\end{bmatrix}.\nonumber
	\end{eqnarray}
\end{small}
The vector $\textbf{x} = [x \ y \ z]^T$ which includes the coordinate of the object that needs to be localized would be: $\textbf{x} = (\textbf{A}^T\textbf{A})^{-1}\textbf{A}^T\textbf{b}$. Further, we can multiply a constant in each row of $\textbf{A}$ and $\textbf{b}$ to give weight according to the channel quality of each receiver, i.e., give weight according to the SNR of the received data. 

\section{Optimal Beacon Placement Problem Formulation} \label{s:Problem_Formulation}
In this section, we derive the trilateration localization error~\ref{ss:GDOP} and introduce a term for quantifying the quality of a beacon configuration. This term is helpful in evaluating and comparing different beacon placement candidates. Then, in~\ref{ss:Problem_Definition}, we elaborate on the problem definition and introduce some terms that we use in formulating and solving the problem.

\subsection{Mathematical Formulation of the Localization Error} \label{ss:GDOP}
A useful metric for quantifying the localization accuracy is the Cramer-Rao Bound (CRB), the lower bound on the location variance that can be achieved using an unbiased location estimator \cite{CMU}. In \cite{CMU}, Niranjini showed that for a $2$D trilateration system with an unbiased estimator, under the assumption that the range measurements are independent and have zero-mean additive Gaussian noise with constant variance $\sigma^2_r$, the CRB variance of the positional error $\sigma^2(r)$ at position $r$, as defined by $\sigma^2(r) = \sigma^2_x(r) + \sigma^2_y(r)$ is given by:
\begin{eqnarray}
	\sigma(r) = \sigma_r \times \sqrt{\frac{N_b}{\sum_{k=1}^{N_b-1}\sum_{j=k+1}^{N_b}F_{kj}}}, \nonumber
\end{eqnarray}
where $N_b$ is the number of beacons, $F_{kj} = |\sin(\theta_k - \theta_j)|$, $\theta_k$ is the angle between $b_k$ and $r$, and $b_k$ is the $k$-th beacon.

This shows that the localization error is a result of the multiplication of the ranging measurement error with another term. This term is a function of the number of beacons and the angle between beacons and the object to localize. In satellite calculations, this function is called Geometric Dilution of Precision (GDOP), therefore: $\sigma(r) = \sigma_r \times GDOP$. As CRB is directly proportional to the GDOP, we can consider GDOP as a reasonable guideline to quantify the localization accuracy \cite{CMU, Relative_location_estimation, Cellular_Mobile_Estimation,Cooperative_localization}.

In general, for three-dimensional localization for an object at $(x,y,z)$, we have:

\begin{eqnarray}
GDOP \cdot \sigma_r = \sqrt{Var(x)+Var(y)+Var(z)+Var(c\tau)}, \nonumber
\end{eqnarray}
where $c$ here is the speed of sound and $\tau$ is the receiver's clock offset. Because we have synchronization between the receiver and the transmitters, the timing offset is considered to be zero, so we have:
\begin{eqnarray} \label{GDOP_1}
GDOP = \sqrt{\frac{\sigma^2_x+\sigma^2_y+\sigma^2_z}{\sigma^2_r}}.
\end{eqnarray}

Let $(x, y, z)$ denote the position of the ultrasound receiver on-board the drone. Let $(x_i, y_i, z_i)$ represent the positions for each of the ultrasound beacons in the room. The drone range to each beacon is calculated from the following:
\begin{eqnarray} \label{r}
r_i = \sqrt{(x-x_i)^2 + (y-y_i)^2 + (z-z_i)^2}.
\end{eqnarray}
Because of errors in measurement and estimation, the amount for different $r_i$ is an estimate and that causes errors in the solution of Eq.~\ref{r} for
$(x, y, z)$. To find a relationship between the solution errors and the ranging errors between the drone and each of the ultrasound transmitter beacons in the room, similar to \cite{xDOP_Formulas}, we take the differential of Eq.~\ref{r} and ignore terms beyond first order (Taylor Expansion):
\begin{eqnarray}
\Delta r_i = \frac{\Delta x(x-x_i) + \Delta y(y-y_i) + \Delta z(z-z_i)}{\sqrt{(x-x_i)^2 + (y-y_i)^2 + (z-z_i)^2}} \nonumber \\
= \Delta x \cos \alpha_i + \Delta y \cos \beta_i + \Delta z \cos \gamma_i \nonumber
\end{eqnarray} 
where $\textbf{U}^i = [\cos \alpha_i \ \cos \beta_i \ \cos \gamma_i]^T$ is the unit vector pointing from the receiver to the $i$-th beacon.

Let $\mathbf{\Delta X} = [\Delta x \ \Delta y \ \Delta z]^T$ be the position error vector and $\mathbf{\Delta R} = [\Delta r_1 \cdots \Delta r_n]^T$ be the target range error vector. Then we can define matrix $\textbf{C}$ as:
\begin{eqnarray}
\textbf{C} & = & \begin{bmatrix}
c^1_1 & c^1_2 & c^1_3 \\
\vdots     & \vdots     & \vdots\\
c^n_1 & c^n_2 & c^n_3 \\
\end{bmatrix}\nonumber
\end{eqnarray}

where $[c^i_1 \ c^i_2 \ c^i_3] = [\cos \alpha_i \cos \beta_i \cos \gamma_i]$. Now we can write $\mathbf{\Delta R} = \textbf{C} \mathbf{\Delta X}$ and then we have $\mathbf{\Delta X} = (\textbf{C}^T\textbf{C})^{-1} \textbf{C}^T \mathbf{\Delta R}$. We know that
\begin{eqnarray} \label{GDOP_2}
\textbf{Cov}(\mathbf{\Delta X}) = \textbf{E}(\mathbf{\Delta X}\mathbf{\Delta X}^T) =   
\begin{bmatrix}
\sigma^2_x & \sigma_{xy} & \sigma_{xz} \\
\sigma_{yx} & \sigma^2_y & \sigma_{yz}  \\
\sigma_{zx} & \sigma_{zy} & \sigma^2_z  \\
\end{bmatrix}. 
\end{eqnarray}
If we assume that Var($r_i$) = $\sigma^2_r$ and that the errors $\Delta r_i$ are uncorrelated, then
\begin{eqnarray} \label{GDOP_3}
\textbf{E}(\mathbf{\Delta X}\mathbf{\Delta X}^T) = \textbf{E}(((\textbf{C}^T\textbf{C})^{-1} \textbf{C}^T \mathbf{\Delta R})((\textbf{C}^T\textbf{C})^{-1} \textbf{C}^T \mathbf{\Delta R})^T) \nonumber \\
= (\textbf{C}^T\textbf{C})^{-1} \textbf{C}^T \textbf{E}(\mathbf{\Delta R}\mathbf{\Delta R}^T) ((\textbf{C}^T\textbf{C})^{-1} \textbf{C}^T)^T \nonumber \\
= (\textbf{C}^T\textbf{C})^{-1} \textbf{C}^T \textbf{C} (\textbf{C}\textbf{C}^T)^{-1} \sigma^2_r = (\textbf{C}^T\textbf{C})^{-1} \sigma^2_r. \nonumber
\end{eqnarray}
Eq.~\ref{GDOP_1}, Eq.~\ref{GDOP_2} and the above result show that the diagonal elements of the $(\textbf{C}^T\textbf{C})^{-1}$ can be used to calculate the GDOP. 

In summary, We showed that localization error comes from two primary sources. The first source is the ranging measurement error caused by the measurement device precision, quality of the received signal, and multi-path. The second source of error is due to the relative geometry between the transmitters and the receiver. The latter part is known as GDOP. Because we have the same ranging measurement error originating from the measurement devices and steady environmental conditions, GDOP can be used as a good measure to quantify the quality for different beacon placements. Table~\ref{table:GDOP} shows the assessment of the GDOP values that affect the accuracy of the localization due to the geometry of beacons.

\begin{table}
	\caption{Evaluation of GDOP Values}
	\label{table:GDOP}
\begin{center}
	\begin{tabular}{ |c | c | }
		\hline
		\textbf{GDOP Values} & \textbf{Evaluation of the geometry of the beacons} \\
		\hline
		$<1$ & Measurements error or redundancy \\
		\hline
		$1$ & Ideal \\
		\hline
		$1-2$ & Very Good  \\
		\hline
		$2-5$ & Good \\
		\hline
		$5-10$ & Medium \\
		\hline
		$10-20$ & Sufficient  \\
		\hline
		$>20$ & Bad \\
		\hline
	\end{tabular}
\end{center}
\end{table}

\subsection{Problem Definition}\label{ss:Problem_Definition}

We are proposing an optimal beacon placement for drone localization in indoor environments. The primary goal is to find the optimal number of ultrasound beacons to achieve full airspace coverage. This means that a drone has to have access to at least four beacons at each point during its flight. Moreover, we want to increase the localization accuracy and mitigate the error induced by the relative geometry between the ultrasound beacons and the drone. Therefore, the proposed optimization algorithm has an additional constraint on the GDOP value at each flying point rejecting configurations with high GDOP values.

The optimal beacon placement for single static target localization in two dimensions is well understood. However, the optimal placement for multiple target locations, a target trace, a target area, and a mobile target trajectory in a defined area are still open problems~\cite{CMU_paper}. Moreover, finding an optimal beacon placement configuration for indoor localization, both minimizing the number of beacons and the localization error at any given position, is a well-established NP-Hard problem~\cite{Efficient_Beacon_Placement, CMU_paper, Novel_Beacon_Placement, BLE_Localization_Precision_Limits}. 

To the best of our knowledge, this is the first work that aims to tackle both the minimization of the number of beacons and the minimization of relative geometry localization error due to the beacon placement concurrently. Our approach considers the errors induced by lack of access to at least four beacons, bad signal reception from some of the beacons, and multi-path to some extent. We perform the latter by finding the number so that all the points have a clear line of sight to at least four beacons. This is done while we attempt to find the minimum number of beacons so that the drone has access to at least four beacons at each point during its flight. A secondary optimization goal is to minimize the error induced due to the relative geometry boosting localization accuracy.

We revisited the Art Gallery problem, a well-established NP-Hard problem in our approach. In this problem, we want to calculate the minimum guards required to fully cover an art gallery so that each point in the gallery is covered by at least one guard. Our problem is similar to the art gallery problem, except that each point needs to have access to at least four beacons, not just one. In addition, our problem is defined in a three-dimensional space, not just two dimensions. The $k$-connectivity problem is to find an arrangement such that each element has access to at least $k$ anchor nodes. Our problem is similar to a combination of the art gallery problem and the $k$-connectivity problem with $k=4$. 

More specifically, minimizing the number of beacons required to fully cover an area is an NP-hard MIP (Mixed Integer Programming) problem. Thus, we formulate a modified version of the MIP problem that includes constraints to also perform optimal placement of beacons, as shown below.

\begin{eqnarray}
\min& \sum_{i=1}^{n} b_i \nonumber \\
s.t.& \sum_{i=1}^{n} bc_{ij} \geq k, \forall j \in D; \nonumber \\
    & GDOP_{avg} \leq g; \nonumber
\end{eqnarray}
where $b_i$ is the value for $i$-th beacons, which equals $1$ if the $i$-th beacon is selected and equals $0$ otherwise. All the beacons are selected from the beacon domain, namely set $B$, which contains all the acceptable locations for beacons in the room. The entire ceiling and top half of all side walls are acceptable candidates for the beacon locations. $n$ is the number of all acceptable beacons available in set $B$. $bc_{ij}$ is the element located at the $i$-th row and the $j$-th column of the connectivity matrix $(\textbf{BC})$ and it is equal to $1$ if the $j$-th point in the drone domain is accessible to the $i$-th beacon from beacon domain and equals $0$ otherwise. The drone domain, set $D$, is a subspace of the room in which the drone is allowed to fly. $k$ is the connectivity number. For our problem $k=4$ because when using 3D trilateration, the distance between the object and at least four beacons is required. $GDOP_{avg}$ is the average of all $GDOP$ values over all the points in the set $D$, and $g$ denotes the average $GDOP$ threshold. We choose the value for $g$ based on Table~\ref{table:GDOP}. 

The first constraint guarantees that each point $j$ in the set $D$ has access to at least four beacons. This constraint guarantees that the final proposed beacon placement can fully cover every point in the set $D$ with at least four beacons. The second constraint guarantees that the final proposed beacon configuration has the optimal placement in order to minimize the geometry-induced localization error for each point in the set $D$.
 
For $GDOP$ calculation, as we discussed in the previous part, if each measurement has the same uncertainty with zero mean and unit variance and they are uncorrelated from each other, then the aforementioned $GDOP$ in the above steps can be derived from the diagonal elements of the matrix $\textbf{Q}$ as follows.
\begin{eqnarray}
\textbf{Q} = (\textbf{C}^T\textbf{C})^{-1} = 
\begin{bmatrix}
\sigma^2_x & \sigma_{xy} & \sigma_{xz} \\
\sigma_{xy} & \sigma^2_y & \sigma_{yz} \\
\sigma_{xz} & \sigma_{yz} & \sigma^2_z  \\
\end{bmatrix}, \nonumber 
\end{eqnarray}
where $GDOP = \sqrt{\sigma^2_x+\sigma^2_y+\sigma^2_z}$, and
\begin{eqnarray}
\textbf{C} =   
\begin{bmatrix}
\frac{x_{1}-x}{r_1} & \frac{y_{1}-y}{r_1} & \frac{z_{1}-z}{r_1} \\
\frac{x_{2}-x}{r_2} & \frac{y_{2}-y}{r_2} & \frac{z_{2}-z}{r_2} \\
\frac{x_{3}-x}{r_3} & \frac{y_{3}-y}{r_3} & \frac{z_{3}-z}{r_3} \\
\frac{x_{4}-x}{r_4} & \frac{y_{4}-y}{r_4} & \frac{z_{4}-z}{r_4} \\
\end{bmatrix}, \nonumber 
\end{eqnarray}
where $(x,y,z)$ is the drone's position, $(x_i,y_i,z_i)$ is the location coordinate of the $i$-th ultrasound beacon, and $r_i$ represents the distance between the drone and the $i$-th ultrasound beacon.

\section{Optimal Beacon Placement Algorithm Design}\label{s:Algorithm_Design}
We developed a greedy algorithm that is based on the class of Evolutionary Algorithms (EAs) to find the optimal placement for beacons. Initially, we identify the minimum number of beacons required to cover the entire domain D with at least four-connectivity. This means that each point in the Domain D has access to at least four beacons. After finding this number, the algorithm will find the optimal placement for these beacons to keep the $GDOP_{avg}$ below the threshold $g$. 

Set $B'$ is a subset of $B$ which is initially empty $(B' = 0)$. At each step of our proposed EA program, one beacon will be added to this set. The goal is to find a $B'$ with the smallest number of elements. At each step of our EA program, $P$ number of random locations are generated (also called individual). Each individual is associated with one set of beacon placements located at a random position selected from the set of all possible locations for beacons in the room (set $B$). Set $B$ consists of locations covering the entire ceiling and top half part of all the walls. Set $B$ was chosen based on all possible trajectories which drone may fly (set $D$), which is the top half of the entire indoor environment. We chose $P = 250$ with the following considerations. For the random generation of each beacon, we generate one on the ceiling and one for each of the walls. This will guarantee that our random generation has an even distribution over the entire set $B$, and at each one of them, we generate $50$ random beacons. There is an initial "seed" of beacons to start with that depends on the size of the room. In our example, we have tested the algorithm with a larger number than $50$, and they did not improve the final performance of the algorithm, only increasing the computational time. Similarly, smaller than $50$ offer faster convergence, but they might miss solutions that are viable. Moreover, when this number is $50$, at the final generation, it provides more configurations with full coverage, which can be fed to the next step of finding the configuration with the lowest $GDOP_{avg}$ and this is helpful to expedite the process in the second stage. However, the selection of the initial "seed" for the number of random generated beacons at any of those allowed boundaries of the room (ceiling and top half of all walls) can be easily selected as 4 times the theoretical minimum to achieve fast convergence without loss of beacon placement solutions that achieve optimal placement.

As a next step, we sort the beacon placement solutions based on our fitness (cost) function. Choosing the proper cost function is the most important part of our proposed EA program. This is because the quality and quantity of the produced solutions are directly associated with the proper selections of the fitness function. Thus, our proposed fitness function consideration has three major components: (i) select placements with the maximum coverage at each step. For instance, in the very first step where $B'$ is empty and the first beacon is about to be chosen, the fitness function tries to find the beacon that has most of its effective coverage range in the required area while avoiding placements that provide coverage for locations outside of $D$. In the $k$-th step of the EA program, this first consideration guarantees the selected beacons are the ones that together provide the maximum coverage for the domain $D$. (ii) At each step, the cost function needs to select beacon placements that together, they provide four-connectivity coverage for domain $D$ in its entirety or more than a threshold parameter selected by the user. (iii) The third consideration is to pick beacon placement that offers placement configurations with $GDOP_{avg}$ below the threshold ($g$) for the entire drone space.

All the aforementioned considerations and constraints are incorporated in our cost function and applied through all the selection steps. Some of the constraints are more important for earlier steps. For example, in earlier steps, the first consideration plays a more important role. In middle generations, after the entire domain $D$ reaches to at least one-connectivity, both the first and second considerations become more dominant. Finally, in the latest generations (after the entire domain $D$ reaches the four-connectivity constraint), then the importance of the third consideration, which is keeping the $GDOP_{avg}$ below a certain threshold $(g)$, is manifested. After sorting the individuals based on the fitness function, at each step (generation), we pick the first $C$ candidate individuals as the selected parents passed on to the next generation and we kill the rest. For our experiments, we set $C=5$.

The $C$ selected candidate individuals from the previous generation become the initial configuration in the new generation. Next, the performance of the new randomly generated individual beacons is tested. The iterations are similar to the very first step where we have $C$ groups and generate fifty random individuals for each group, then sort these $250$ configurations based on their fitness and choose the first $C$ to go as parents to the next step.

The program terminated when the following criteria were achieved: (i) all the points in the domain $D$ has access to at least four beacons, (ii) and the final configuration has a $GDOP_{avg}$ below a threshold $g$. 

\begin{algorithm}[t]
	\caption{4-Connectivity Optimal Beacon Placement}\label{Alg:Beacon_Placement} 
	\begin{algorithmic}[1]
		\newcommand\algorithmicinput{\textbf{Input:}}
		\newcommand\INPUT{\item[\algorithmicinput]}
		\INPUT Drone domain (D), Beacon domain (B), Beacon Range (R), K-connectivity (K).
		\newcommand\algorithmicoutput{\textbf{Output:}}
		\newcommand\OUTPUT{\item[\algorithmicoutput]}
		\OUTPUT Beacon placement configuration with minimum number of beacons and 4-connectivity full coverage.
		\newcommand\algorithmicinitialization{\textbf{Initialization:}}
		\newcommand\INITIALIZATION{\item[\algorithmicinitialization]}
		\INITIALIZATION
		\FOR {$i = 1$ to $i = number \ of \ individuals \ (P)$}
		\STATE Generate one beacon at random position $(x,y,z) \in B$;
		\STATE Calculate the overall coverage provided by this beacon (fitness).
		\ENDFOR
		\STATE Sort all of the individuals ($P$ sets of one beacon) with respect to their fitness from the highest to the lowest;
		\STATE Select the first $C$ of them (the $C$ best of them according to their fitness) and eliminate the rest $P - C$ (Evolution chooses the best as parents for the next generation and kills the rest); 
		\STATE Update $P_1, \cdots , P_{C}$ with these survivor individuals which are going to be used in the next generation.
		\newcommand\algorithmicoptimization{\textbf{Optimization Framework:}}
		\newcommand\OPTIMIZATION{\item[\algorithmicoptimization]}
		\OPTIMIZATION
		\FOR {$k = 1$ to $i = 4 \ (K-connectivity = 4)$}
		\WHILE{$STOP == 0$}
		\FOR {$i = 1$ to $i = P/C$}
		\STATE For each of the selected individuals from previous generation $(P_1, \cdots , P_{C})$, generate one beacon at random position $(x,y,z) \in B$ and add it to them;
		\STATE Calculate the $k$ fitness (maximum coverage with respect to each point has access to $k$ beacon) for each of these individuals.
		\ENDFOR
		\STATE Sort all of these new individuals ($P$ sets of some beacons) with respect to their $k$ fitness from the highest to the lowest;
		\STATE Select the first $C$ of them (the $C$ best of them according to their fitness) and eliminate the rest $P - C$ (Evolution chooses the best as parents for the next generation and kills the rest); 
		\STATE Update $P_1 \cdots P_{C}$ with these individuals which are going to be be used in the next generation;
		\IF {at least one of the $\{P_1, \cdots , P_{C}\}$ has a $k$ connectivity coverage}
		\STATE $STOP = 1$
		\ENDIF
		\ENDWHILE
		\ENDFOR	
	\end{algorithmic}
\end{algorithm}

\begin{algorithm}
	\caption{GDOP Optimal Beacon Placement}\label{Alg:GDOP}
	\begin{algorithmic}[1]
		\newcommand\algorithmicinput{\textbf{Input:}}
		\newcommand\INPUT{\item[\algorithmicinput]}
		\INPUT Drone domain (D), Beacon domain (B), Beacon Range (R), K-connectivity (K), $GDOP_{avg}$ threshold ($g$), The number of required beacons ($N$) obtained from the other Algorithm, The final set $\{P_1, \cdots , P_{C}\}$ obtained from the other Algorithm.
		\newcommand\algorithmicoutput{\textbf{Output:}}
		\newcommand\OUTPUT{\item[\algorithmicoutput]}
		\OUTPUT Optimal Placement for N beacons that provides four-connectivity and keeps the $GDOP_{avg}$ below the threshold $g$ over the entire domain $D$.
		\newcommand\algorithmicinitialization{\textbf{Initialization:}}
		\newcommand\INITIALIZATION{\item[\algorithmicinitialization]}
		\INITIALIZATION Individuals in this algorithm are sets of $N$ beacons. The initial individuals are $\{P_1, \cdots , P_{C}\}$ from the last algorithm.
		\newcommand\algorithmicoptimization{\textbf{Optimization Framework:}}
		\newcommand\OPTIMIZATION{\item[\algorithmicoptimization]}
		\OPTIMIZATION
		\WHILE {$four-connectivity == 0$}
		\WHILE{$GDOP_{avg} > g$}
		\FOR {$i = 1$ to $i = C$}
		\FOR {all the points $(x,y,z) \in D$}
		\STATE Calculate the $GDOP$ at point $(x,y,z)$ from $(C^TC)^{-1}$;
		\ENDFOR
		\STATE Calculate the $GDOP_{avg}$ over all the points in $D$.
		\ENDFOR
		\STATE Sort individuals based on the calculated $GDOP_{avg}$ (fitness) from the lowest to highest (lower is better);
		\STATE Select the individuals with better fitness as Parents;
		\STATE Crossover each two adjacent parents and make a new offspring individual;
		\STATE Kill the worst ones to keep having $C$ individuals;
		\STATE Update $P_1, \cdots , P_{C}$ with these new survivor individuals.
		\ENDWHILE
		\IF {at least one of the $\{P_1, \cdots , P_{C}\}$ has the full four-connectivity coverage over entire set $D$}
		\STATE $four-connectivity = 1$
		\ENDIF
		\ENDWHILE
	\end{algorithmic}
\end{algorithm}

\section{Simulation Setup \& Evaluation} \label{s:Results}
In this section, we describe our simulation setup and provide a comparative analysis of the expected theoretical outcomes and our simulation results, including parameters and methodology.
\subsection{Simulation Setup}
We use MATLAB software to simulate and evaluate OPTILOD's performance. The software runs on a Dell OptiPlex~$7080$ Desktop. Our algorithm~\ref{Alg:Beacon_Placement} has three major sections: (i) main algorithm, (ii) functions block, (iii) $GDOP_{avg}$ constrained configuration algorithm.

The main algorithm section calculates the minimum number of beacons to achieve four-beacon coverage. The algorithm begins by selecting a random beacon from the beacon space (set $B$) and examining the range provided by this beacon while considering all of the ultrasound sensors' and floor plan's limitations. Each step aims to select the beacon that offers the maximum coverage for that k-connectivity step when combined with the other available beacons. This section produces sets of beacon configurations, some of which provide full coverage for the entire drone space (i.e., 100\% coverage). The rest provide coverage that is greater than the desired threshold (e.g., 97\% coverage). Finally, the algorithm ranks the sets of beacon configurations based on their maximum coverage, with the final output being those with 100\% full coverage.


In terms of function blocks, we have \textit{coverage.m}, \textit{fitness.m}, and \textit{totalcoverage.m} as our functions. The first one, \textit{coverage}, shows the coverage provided by one beacon placed at position $(x,y,z) \in B$. This coverage is based on all the limitations for the ultrasound propagation patterns in addition to the restrictions and boundaries of the desired floor plan. In the \textit{fitness} function, we have four sub-blocks each of them responsible for k-connectivity where $k$ goes from $1$ to $4$. In the main code, we use 1-connectivity coverage first until we achieve the 1-connectivity coverage for all points (i.e., access to one beacon). We then move to achieve 2-connectivity coverage and so on until we achieve 4-connectivity coverage. The last function, \textit{totalcoverage} is used to check the total coverage provided by all the beacons in that configuration. When the total coverage is the entire drone space (set $D$), then the main code for the first stage stops and the second stage, which is $GDOP_{avg}$ constrained code, starts.

The $GDOP_{avg}$ constrained configuration code selects the configurations with $GDOP_{avg} < g$ among all the candidate configuration from the last step. We can then rank all the configurations that achieve 100\% coverage or a configurable threshold based on the user input. This allows for configurations with less than $100\%$ coverage but above a threshold (i.e., $97\%$) to be considered as solutions creating a trade-off between localization error minimization and room coverage that can be leveraged to our advantage. Such a trade-off is important for our solution space. Solutions with $100\%$ coverage can result in significantly more localization error than ones with lower coverage but better relative geometry.

\subsection{Results \& Comparative Analysis}
This section discusses our simulations' results and contrasts them with the theoretically expected outcomes. To provide reliable test results, we run multiple tests with different floor plans evaluating the performance of our proposed algorithm under various conditions. We split the problem into optimization goals verifying each goal separately against the theoretical optimal. The first goal for our optimization program is to identify the minimum number of beacons required to achieve full area cover. The optimal placement would guarantee that the set $D$ of points representing the drone airspace have access to at least four beacons, a condition necessary to achieve self-localization. To validate our approach, we compare the number of needed beacons derived from our proposed algorithm with the lower bound on the minimum number of beacons that can be obtained theoretically. 

Moreover, OPTILOD considers the limitations of the ultrasonic sensors: propagation pattern, range of work, and angle of the propagation beam. In this paper, whenever we mention the sensors or beacons, we mean the sensor array, where each sensor array includes $6$ ultrasound sensors. The $6$ sensor array as one beacon is necessary to compensate for the narrow-beam propagation pattern of the ultrasound sensors. We will walk the reader through a simple scenario consisting of a cubic floor plan. One such example is an area of $3$~m $\times$ $3$~m $\times$ $4$~m. In this example, we can actually achieve the lower theoretical bound on the minimum number of beacons which is $4$. Our proposed optimization program calculates the minimum number to be $4$ as well, which is exactly equal to the same as the theory. Moreover, our algorithm achieves this number very quickly. In addition, our approach was able to provide $250$ sets of $4$ sensor arrays with both full and partial (configurable) coverage. Almost one-fifth of these configurations achieved $100\%$ coverage and the majority of the rest achieved at least $96\%$. This flexibility is important because it allows us to offer more alternative configurations when trying to achieve our secondary optimization goal of reducing the localization error.

\begin{figure}[t]
	\centering
	\includegraphics[width=0.49\textwidth]{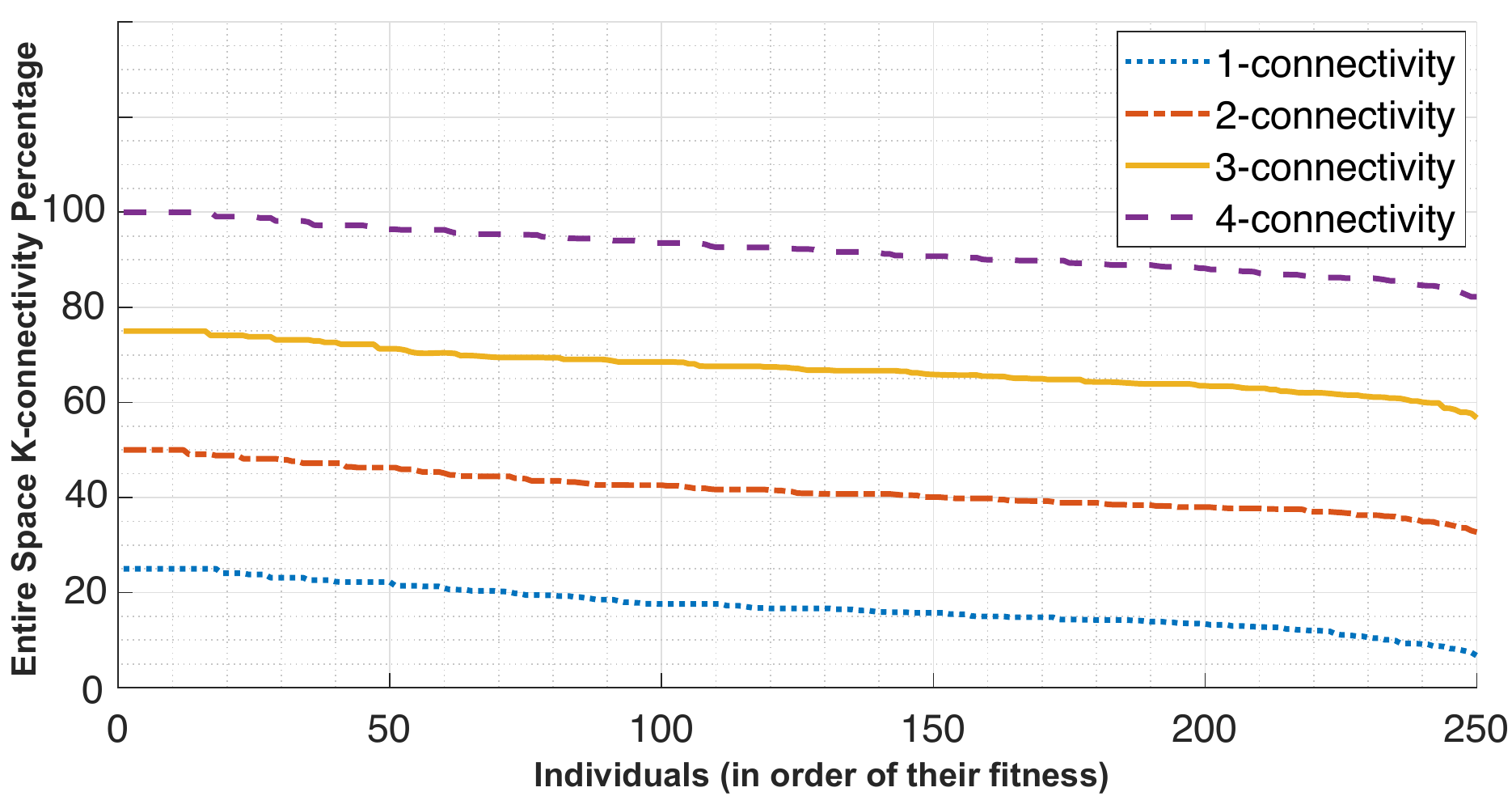}
	\caption{Percentage of the connectivity achieved for the entire drone space after finishing each of the k-connectivity steps in the first stage of OPTILOD.}
	\label{fig:K_Connectivity_Percentage}
\end{figure}

Figure~\ref{fig:K_Connectivity_Percentage} depicts the percentage of the k-connectivity achieved for covering the entire drone space. As seen in this figure, after finishing the first stage of the k-connectivity algorithm, the last generation has the maximum value $100\%$ coverage (shown as $25\%$ as it is only 1-coverage as opposed to 4-coverage we need) for the fittest population of beacon configurations. The amount of viable beacon placements decreases as we increase the connectivity requirements. This means that after finishing the 1-connectivity algorithm, the scheme outputs $250$ beacon placement configurations. As the algorithm progresses, the fitness function adjusts to coverage for 2-connectivity, 3-connectivity, and finally 4-connectivity. At each step, the beacon placements that offer full coverage (our population), are selected as the parents for the next generation. For instance, after the first step, the first $10$ beacon configurations among the total $250$ have the $100\%$ 1-connectivity full coverage for the entire drone space and are selected as candidates for 2-connectivity and so on.

\begin{figure}[t]
	\centering
	\includegraphics[width=0.5\textwidth]{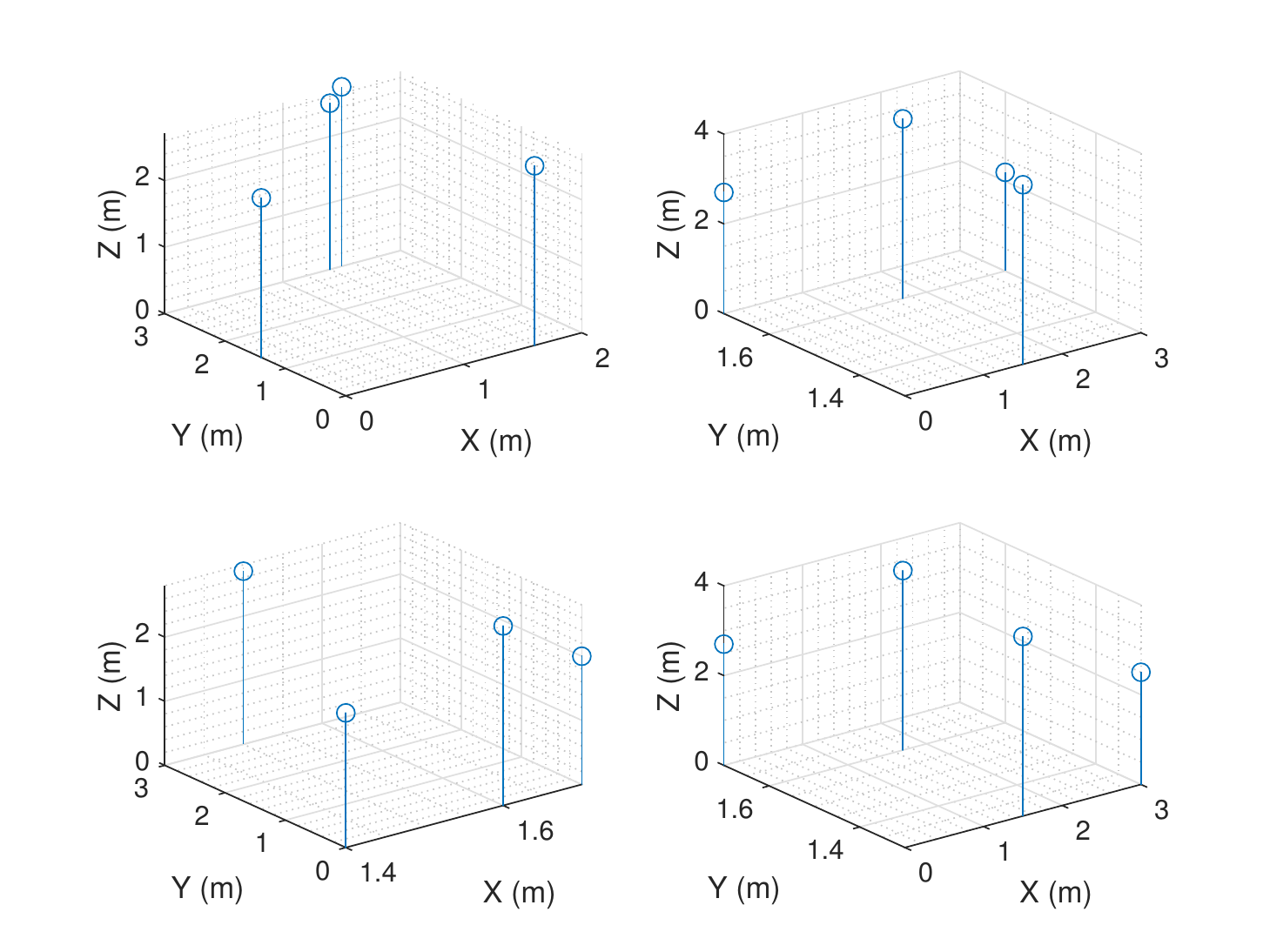}
	\caption{Alternative placement candidates offered by OPTILOD for a simple scenario: room with dimensions $3$ $\times$ $3$ $\times$ $4$. The blue circles represent ultrasound beacons. These placements offer both full four beacon connectivity coverage and a "sufficient" average $GDOP$.}
	\label{fig:334_Final_Placement}
\end{figure}

For the second stage of OPTILOD, the beacon configurations produced by the first step are inserted as input into the Algorithm~\ref{Alg:GDOP}. Notice that, for step two, we aim to reduce the localization error due to the relative geometry between the drone and the beacons. Therefore, our stopping condition is at least one placement configuration with an average $GDOP$ below the average threshold $g$ from Table~\ref{table:GDOP} over all the points in the drone space (set $D$). This ensures that our final beacon placement configuration not only provides the full coverage, but it also induces the minimum localization error for the threshold ($g$). The outcome of this step is the beacon configuration which has the full 4-connectivity coverage and also induces low localization error. Figure~\ref{fig:334_Final_Placement} shows the outcome configurations after applying both steps representing four alternative beacon configurations where all of them have the full connectivity and average $GDOP$ requirements. 

\begin{figure}[t]
	\centering
	\includegraphics[width=0.49\textwidth]{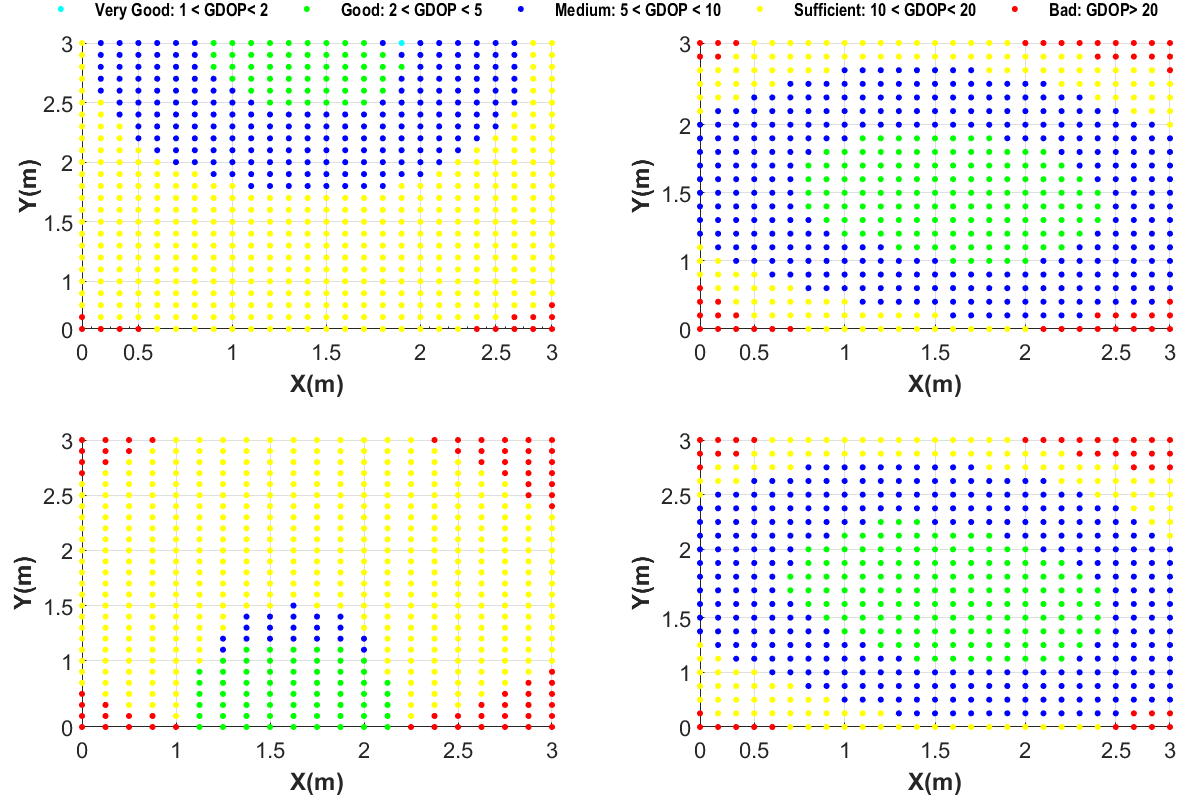}
	\caption{$GDOP$ representation of the entire drone space corresponding to the four beacon configurations in Figure~\ref{fig:334_Final_Placement}. Notice that while all the configurations provide 100\% coverage, not all configurations are equal when it comes to GDOP. The red dots in the plots indicate "bad" GDOP are concentrated in the corners of the area under consideration with the bottom right placement being the best (on average). Drone flight paths are designed to avoid room corners as they are collision prone. Thus, in practice, OPTILOD provides good to very good GDOP for all usable navigation paths.}
	\label{fig:334_GDOP_Representation}
\end{figure}

Figure~\ref{fig:334_GDOP_Representation} illustrates the $GDOP$ values for each points in the drone space corresponding to the four alternative beacon configurations presented in Figure~\ref{fig:334_Final_Placement}. In Figure~\ref{fig:334_GDOP_Representation}, we calculate the $GDOP$ value for each point in the drone space and then project it on two dimensions: $X$ and $Y$. This is to simplify understanding the plot; otherwise, the original figure was in three dimensions. As is seen in this figure, for all the four alternative configurations, the majority of the drone space has a $GDOP$ value between $10$ to $20$ and this is due to the fact that we set $g$ in a way to achieve only "sufficient" average $GDOP$. Moreover, based on the $GDOP$ representation of these four beacon configurations, we have the freedom to pick the one that provides a better $GDOP_{avg}$ if we know the more probable flight trajectories in advance. 

\begin{figure}[t]
	\centering
	\includegraphics[width=0.49\textwidth]{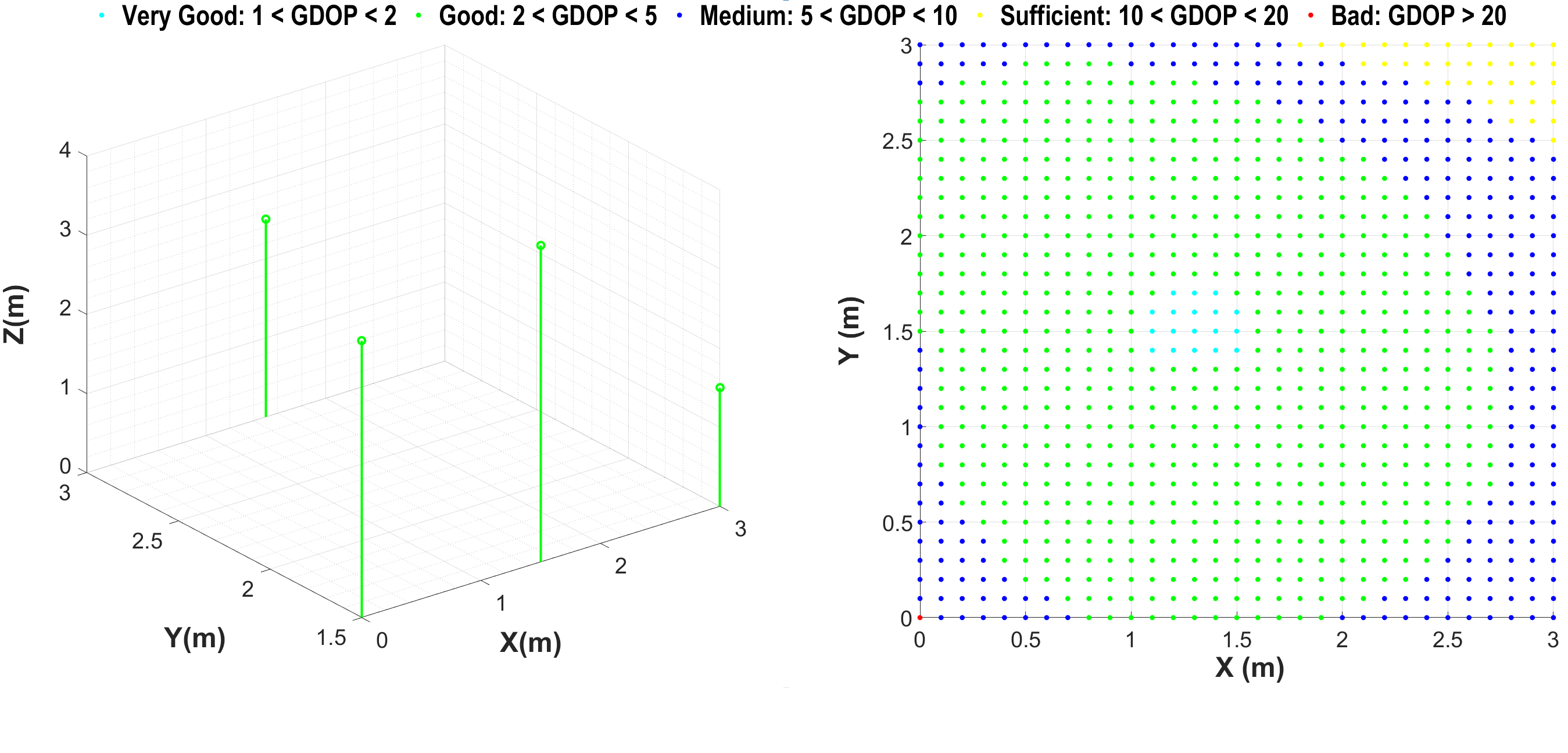}
	\caption{Beacon configuration and its corresponding $GDOP$ representation for a room with dimensions $3$~m $\times$ $3$~m $\times$ $4$~m. Left: The beacon configuration with the green circles as the ultrasound sensor arrays. This configuration provides 4-connectivity for 96\% of the room. However, it has a much better performance in terms of localization error. Right: The calculated $GDOP$ values for the entire room with majority having the value less than $5$. The $GDOP_{avg}$ is 2.8 (Good).}
	\label{fig:334_lower_GDOP_Representation}
\end{figure}

To achieve a beacon configuration with a smaller $GDOP_{avg}$, there is a trade-off between the time and coverage: we can either have a full 4-connectivity coverage and a smaller $GDOP_{avg}$ but a longer process time or we can achieve smaller $GDOP_{avg}$ in a very short processing time but if we accept the 4-connectivity coverage for $96\%$ of the drone space. Figure~\ref{fig:334_lower_GDOP_Representation} showcase this fact better. As is seen in this figure, the provided beacon configuration has the $4$-connectivity for $96\%$ of the drone space, but instead, the $GDOP$ representation shows that the majority of the room has a value between $2$ to $5$ for the $GDOP$ and the $GDOP_{avg}$ is "good" over the entire drone space which means much better localization accuracy.

Another example among so many in which we evaluated OPTILOD is an area of $5$~m $\times$ $5$~m $\times$ $4$~m. For this example, after considering all the limitations, the lower bound on the minimum number found by theory is $16$ sensor arrays, whereas explained before, each sensor array includes $6$ ultrasound beacons. Our proposed optimization program calculates the minimum number to be $17$, which is just one sensor array more than the actual lower bound, still within the expected error. Moreover, our proposed algorithm managed to solve this problem $6$ times faster than the comparative greedy evolutionary algorithm that can only solve the problem for full coverage configurations with $16$ sensor arrays. Contrary to that, our approach was able to provide $250$ sets of $17$ sensor arrays with both full and partial (configurable) coverage. Almost one-fifth of these configurations achieved $100\%$ coverage, with the rest achieving at least $97\%$. This flexibility is essential because it allows us to offer more alternative configurations when we try to achieve our secondary optimization goal.

We bootstrap the second stage of our solution using the beacon configurations produced by our first step. The goal here is to find the arrangements with the average $GDOP$ below a threshold $g$ to guarantee lower localization error induced by the relative geometry between the drone and the beacons. 
To showcase the trade-off between the observed minimum number of beacons required for 4-connectivity full coverage over the entire drone space and the threshold $g$, which dictates keeping the average $GDOP$ below a certain amount to provide better localization accuracy, we run OPTILOD for the similar case with a smaller threshold $g$ requirement. This results in the first stage of OPTILOD to find the minimum number of beacons to be $18$ instead of $17$. This number is still very close to the minimum number calculated by theory, but the final beacon configurations have much better localization accuracy. The average $GDOP$ of these placements is almost one-fifth of those with $17$ beacons.

\begin{figure}[t]
	\centering
	\includegraphics[width=0.49\textwidth]{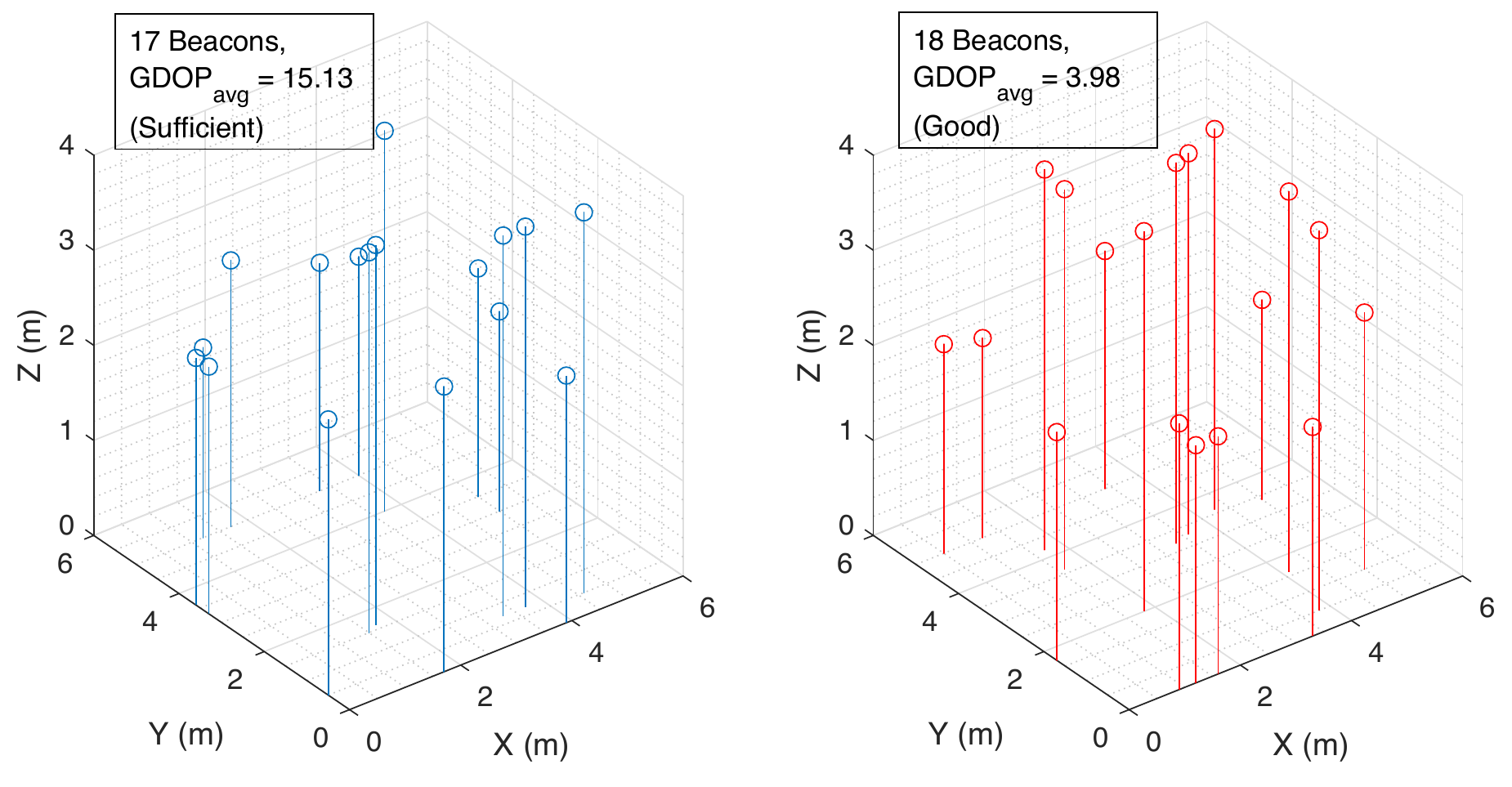}
	\caption{Beacon configuration for a room with dimensions of $5$~m $\times$ $5$~m $\times$ $4$~m. Left: Configuration with $17$ beacons (blue circles) and a "sufficient" $GDOP_{avg}$. Right: Configuration with $17$ beacons (red circles) and a "good" $GDOP_{avg}$, which means lower localization error.}
	\label{fig:554_17vs18}
\end{figure}

Figure~\ref{fig:554_17vs18} showcase the final beacon configuration for a room with dimensions of $5$~m $\times$ $5$~m $\times$ $4$~m. As is seen in this figure, the left side is a configuration with $17$ beacons, and the right side represents a configuration with $18$ beacons. Both of these configurations have the full 4-connectivity coverage for the entire drone space. However, the left one has a "sufficient" $GDOP_{avg}$, and the one in the right provides a "good" $GDOP_{avg}$, which means lower localization error due to the relative geometry between the drone and the beacons. Based on the problem, if the cost of implementation has the first priority, the left one has one beacon less, which makes it cheaper. However, if the localization accuracy is more important than the price of just one beacon, the configuration in the right provides a better solution.



\begin{figure}[t]
	\centering
	\includegraphics[width=0.49\textwidth]{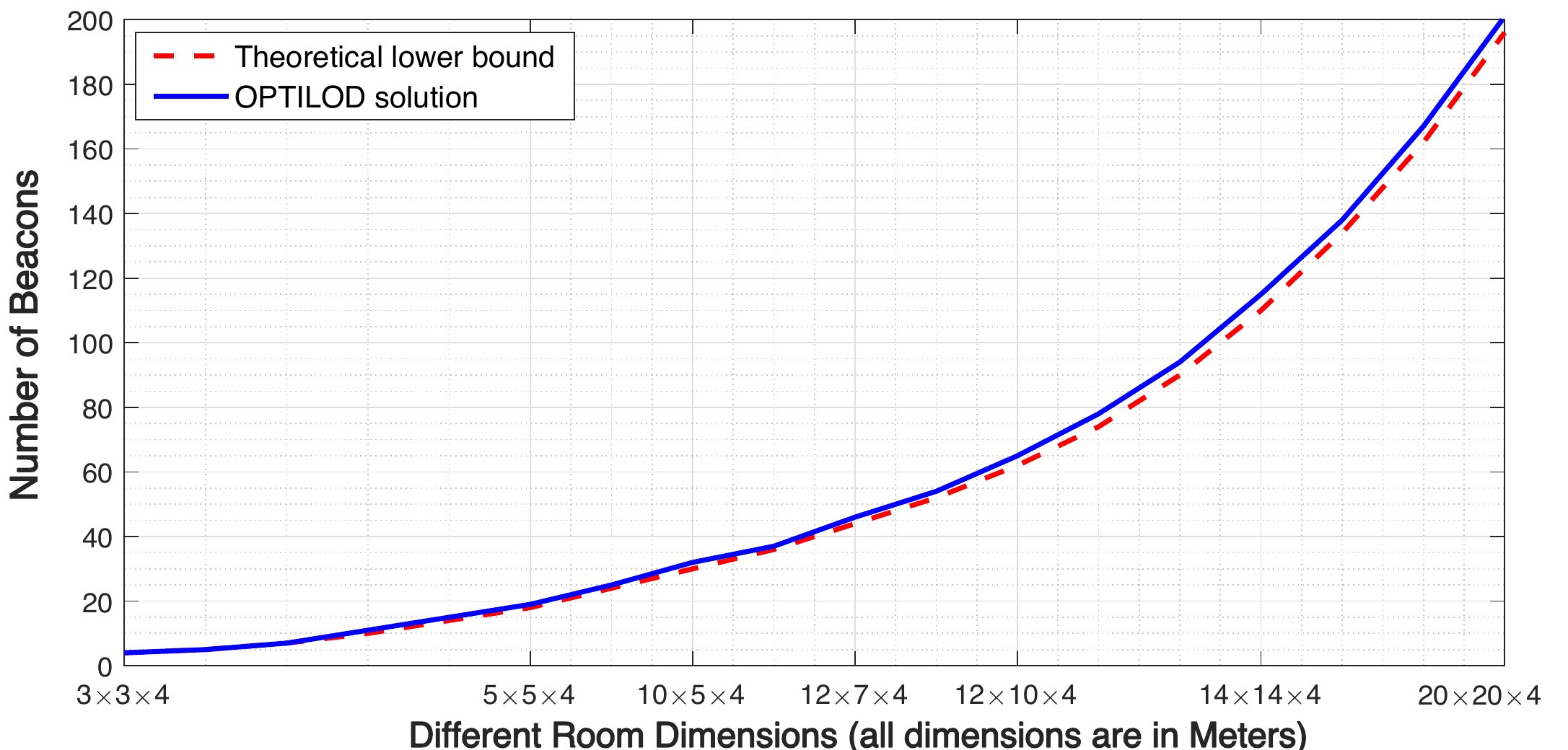}
	\caption{Comparison between the performance of the first stage of OPTILOD  --finding the minimum number of beacons for full 4-connectivity coverage-- with the lower on the minimum number of beacons calculated from theory.}
	\label{fig:THEORYvsOPTILOD}
\end{figure}

As is seen in Figure~\ref{fig:THEORYvsOPTILOD}, OPTILOD manages to perform well in any floor plan regardless of the dimension or other limitation such as the blockage of the signal, multipath, etc. The minimum number of beacons provided by the first stage of OPTILOD is always in very close vicinity of the theoretical lower bound.

We perform our algorithm over multiple other random setups with different room dimensions and floor plan designs. Some of the examples include an office room with $4$~m $\times$ $4$~m $\times$ $4$~m dimensions, a conference room with dimensions of $5$~m $\times$ $5$~m $\times$ $4$~m, and a large hallway with dimensions of $12$~m $\times$ $7$~m $\times$ $4$~m. The final result for all the different scenarios indicates that OPTILOD is capable of achieving the minimum number of the beacons to provide 4-connectivity with the $GDOP_{avg}$ below a desired threshold $(g)$ in a relatively fast process time. The final results are always located reasonably close to the theoretical solution for the number of beacons. More importantly, OPTILOD successfully performs in any floor plan with all the possible items that may block the signal, produce multipath, etc. All of these limitations are considered in the \textit{coverage} function and also by segmentation of the space. For instance, if an item blocks the signal from propagation, we consider that item as an extra wall, make the new floor plan, and solve the problem for the new setup. 

\section{Conclusion} \label{s:Conclusion}
We present  OPTILOD, a framework for confined indoor and underground drone navigation using beacons for self-localization. Our approach works in the absence of GPS and visually impaired environments. OPTILOD depends on a novel optimization algorithm that achieves optimal placement of ultrasound transmitter beacons and, at the same time, reduces localization errors. Thus, a primary design goal of OPTILOD was to identify beacon placement configurations in which the minimum number of ultrasound transmitter beacons are employed for four-beacon coverage at all times. In addition, OPTILOD accomplishes a secondary optimization objective, minimizing localization error caused by the relative geometry between the transmitter beacons and the drone. Our approach is the first to achieve both the minimum number of beacons and their optimal placement for better localization accuracy for indoor drones. We evaluate OPTILOD using extensive simulations for different area sizes and beacon configurations. Our results show that OPTILOD produces beacon placements that are identical or one-beacon more compared to the theoretical optimal bound. Moreover, by calculating the Geometric Dilution of Precision (GDOP) for the beacon placements, we show that OPTILOD produces solutions that have low GDOP (\textit{i.e., low error}). Finally, we demonstrate that the combined optimization problem is tractable even though the original optimization problems independently belong to the NP-Hard Mixed Integer Programming class.


\footnotesize
\bibliographystyle{IEEEtran}
\bibliography{ref_OPTILOD}

\end{document}